# Adversarial Example Generation with Syntactically Controlled Paraphrase Networks

**Mohit Iyyer**[★†‡] **John Wieting**[★♣] **Kevin Gimpel**[♢] **Luke Zettlemoyer**[♠]

Allen Institute of Artificial Intelligence[†] UMass Amherst[‡] Carnegie Mellon University[♣]
Toyota Technological Institute at Chicago[♢] University of Washington[♠]
miyyer@cs.umass.edu jwieting@cs.cmu.edu
kgimpel@ttic.edu lsz@cs.washington.edu

## Abstract

We propose *syntactically controlled paraphrase networks* (SCPNs) and use them to generate adversarial examples. Given a sentence and a target syntactic form (e.g., a constituency parse), SCPNs are trained to produce a paraphrase of the sentence with the desired syntax. We show it is possible to create training data for this task by first doing backtranslation at a very large scale, and then using a parser to label the syntactic transformations that naturally occur during this process. Such data allows us to train a neural encoder-decoder model with extra inputs to specify the target syntax. A combination of automated and human evaluations show that SCPNs generate paraphrases that follow their target specifications without decreasing paraphrase quality when compared to baseline (uncontrolled) paraphrase systems. Furthermore, they are more capable of generating syntactically adversarial examples that both (1) "fool" pretrained models and (2) improve the robustness of these models to syntactic variation when used to augment their training data.

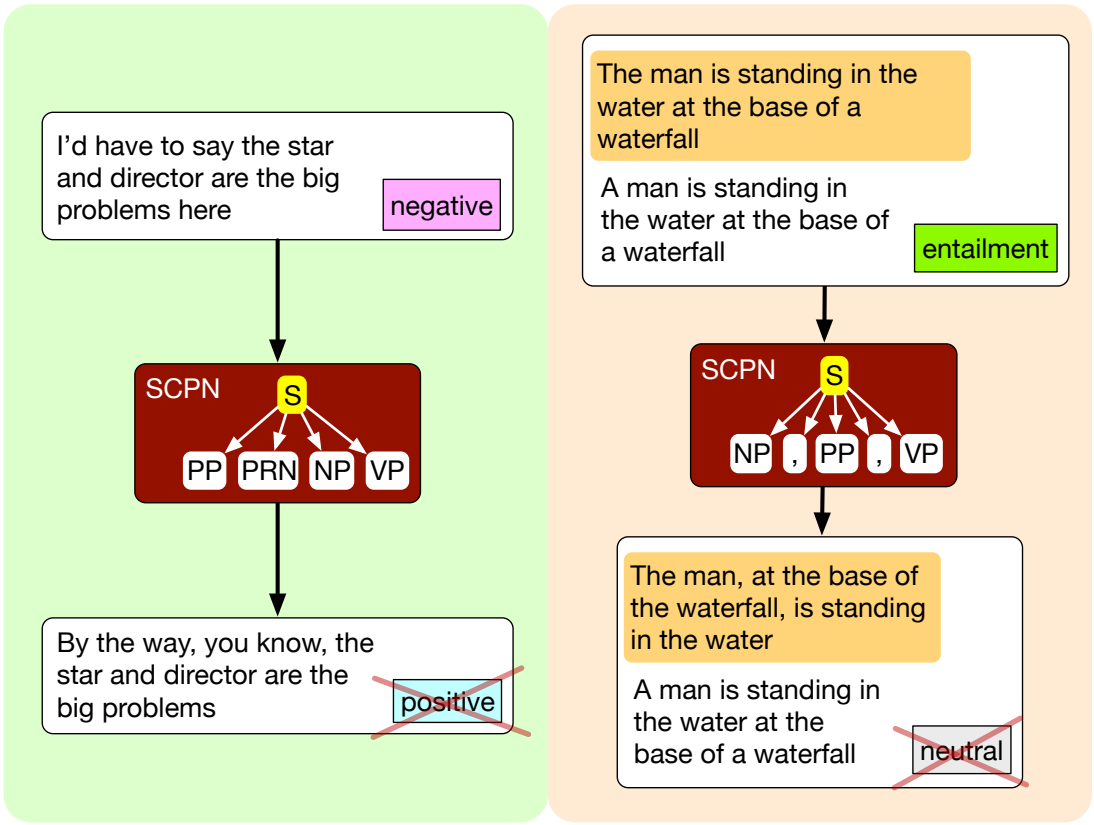

Figure 1: Adversarial examples for sentiment analysis (left) and textual entailment (right) generated by our syntactically controlled paraphrase network (SCPN) according to provided parse templates. In both cases, a pretrained classifier correctly predicts the label of the original sentence but not the corresponding paraphrase.

## 1 Introduction

Natural language processing datasets often suffer from a dearth of linguistic variation, which can hurt the generalization of models trained on them. Recent work has shown it is possible to easily "break" many learned models by evaluating them on *adversarial examples* (Goodfellow et al., 2015), which are generated by manually introducing lexical, pragmatic, and syntactic variation not seen in the training set (Ettinger et al., 2017). Robustness to such adversarial examples can potentially be improved by augmenting the training data, as shown by prior work that introduces rule-based lexical substitutions (Jia and Liang, 2017; Liang et al., 2017). However, more complex transformations, such as generating syntactically adversarial examples, remain an open challenge, as input semantics must be preserved in the face of potentially substantial structural modifications. In this paper, we introduce a new approach for learning to do *syntactically controlled paraphrase generation*: given a sentence and a target syntactic form (e.g., a constituency parse), a system must produce a paraphrase of the sentence whose syntax conforms to the target.

General purpose syntactically controlled paraphrase generation is a challenging task. Approaches that rely on handcrafted rules and grammars, such as the question generation system of McKeown (1983), support only a limited number of syntactic targets. We introduce the first learning approach for this problem, building on the generality of neural encoder-decoder models to support a wide range of transformations. In doing so, we face two new challenges: (1) obtaining a large amount of paraphrase pairs for training, and (2) defining syntactic transformations with which to label these pairs.

★Authors contributed equally.

Since no large-scale dataset of sentential paraphrases exists publicly, we follow Wieting et al. (2017) and automatically generate millions of paraphrase pairs using neural *backtranslation*. Backtranslation naturally injects linguistic variation between the original sentence and its backtranslated counterpart. By running the process at a very large scale and testing for the specific variations we want to produce, we can gather ample input-output pairs for a wide range of phenomena. Our focus is on syntactic transformations, which we define using templates derived from linearized constituency parses (§2). Given such parallel data, we can easily train an encoder-decoder model that takes a sentence and target syntactic template as input, and produces the desired paraphrase.[1]

A combination of automated and human evaluations show that the generated paraphrases almost always follow their target specifications, while paraphrase quality does not significantly deteriorate compared to vanilla neural backtranslation (§4). Our model, the syntactically controlled paraphrase network (SCPN), is capable of generating adversarial examples for sentiment analysis and textual entailment datasets that significantly impact the performance of pretrained models (Figure 1). We also show that augmenting training sets with such examples improves robustness without harming accuracy on the original test sets (§5). Together these results not only establish the first general purpose syntactically controlled paraphrase approach, but also suggest that this general paradigm could be used for controlling many other aspects of the target text.

## 2 Collecting labeled paraphrase pairs

In this section, we describe a general purpose process for gathering and labeling training data for controlled paraphrase generation.

### 2.1 Paraphrase data via backtranslation

Inducing paraphrases from bilingual data has long been an effective method to overcome data limitations. In particular, bilingual pivoting (Bannard and Callison-Burch, 2005) finds quality paraphrases by pivoting through a different language. Mallinson et al. (2017) show that neural machine translation (NMT) systems outperform phrase-based MT on several paraphrase evaluation metrics.

In this paper, we use the PARANMT-50M corpus from Wieting and Gimpel (2017). This corpus consists of over 50 million paraphrases obtained by backtranslating the Czech side of the CzEng (Bojar et al., 2016) parallel corpus. The pretrained Czech-English model used for translation came from the Nematus NMT system (Sennrich et al., 2017). The training data of this system includes four sources: Common Crawl, CzEng 1.6, Europarl, and News Commentary. The CzEng corpus is the largest of these four and was found to have significantly more syntactic diversity than the other data sources (Wieting and Gimpel, 2017).[2]

### 2.2 Automatically labeling paraphrases with syntactic transformations

We need labeled transformations in addition to paraphrase pairs to train a controlled paraphrase model. Manually annotating each of the millions of paraphrase pairs is clearly infeasible. Our key insight is that target transformations can be detected (with some noise) simply by parsing these pairs.[3]

Specifically, we parse the backtranslated paraphrases using the Stanford parser (Manning et al., 2014),[4] which yields a pair of constituency parses $\langle p_1, p_2 \rangle$ for each sentence pair $\langle s_1, s_2 \rangle$, where $s_1$ is the reference English sentence in the CzEng corpus and $s_2$ is its backtranslated counterpart. For syntactically controlled paraphrasing, we assume $s_1$ and $p_2$ are inputs, and the model is trained to produce $s_2$. To overcome learned biases of the NMT system, we also include reversed pairs $\langle s_2, s_1 \rangle$ during training.

#### 2.2.1 Syntactic templates

To provide syntactic control, we linearize the bracketed parse structure without leaf nodes (i.e., tokens). For example, the corresponding linearized parse tree for the sentence "*She drove home.*" is `(S(NP(PRP))(VP(VBD)(NP(NN)))(.))`. A system that requires a complete linearized target parse at test-time is unwieldy; how do we go about choosing the target parse? To simplify test-time usage, we relax the target syntactic form to a parse *template*, which we define as the top two levels of the linearized parse tree (the level immediately below the root along with the root); the prior example's template is `S` → `NP VP`. In the next section, we design models such that users can feed in either parse templates or full parses depending on their desired level of control.

[1]Code, labeled data, and pretrained models available at https://github.com/miyyer/scpn.

[2]Syntactic diversity was measured by the entropy of the top two levels of parse trees in the corpora.

[3]Similar automated filtering could be used to produce data for many other transformations, such as tense changes, point-of-view shifts, and even stylometric pattern differences (Feng et al., 2012). This is an interesting area for future work.

[4]Because of the large dataset size, we use the faster but less accurate shift-reduce parser written by John Bauer.

# 3 Syntactically Controlled Paraphrase Networks

The SCPN encoder-decoder architecture is built from standard neural modules, as we describe in this section.

## 3.1 Neural controlled paraphrase generation

Given a sentential paraphrase pair $\langle s_1, s_2 \rangle$ and a corresponding target syntax tree $p_2$ for $s_2$, we encode $s_1$ using a bidirectional LSTM (Hochreiter and Schmidhuber, 1997), and our decoder is a two-layer LSTM augmented with soft attention over the encoded states (Bahdanau et al., 2014) as well as a copy mechanism (See et al., 2017). Following existing work in NMT (Sennrich et al., 2015), we preprocess $s_1$ and $s_2$ into subword units using byte pair encoding, and we perform decoding using beam search. For all attention computations, we use a bilinear product with a learned parameter matrix $\mathbf{W}$: given vectors $\boldsymbol{u}$ and $\boldsymbol{v}$, we score them by $\boldsymbol{u}^T \mathbf{W} \boldsymbol{v}$.

We incorporate the target syntax $p_2$ into the generation process by modifying the inputs to the decoder. In particular, a standard decoder LSTM receives two inputs at every time step: (1) the embedding $\boldsymbol{w}_{t-1}$ of the ground-truth previous word in $s_2$, and (2) an attention-weighted average $\boldsymbol{a}_t$ of the encoder's hidden states. We additionally provide a representation $\boldsymbol{z}_t$ of the target $p_2$, so at every time step the decoder computes

$$\boldsymbol{h}_t = \text{LSTM}([\boldsymbol{w}_{t-1}; \boldsymbol{a}_t; \boldsymbol{z}_t]). \tag{1}$$

Since we preserve bracketed parse structure, our linearized parses can have hundreds of tokens. Forcing all of the relevant information contained by the parse tree into a single fixed representation (i.e., the last hidden state of an LSTM) is difficult with such large sequences. Intuitively, we want the decoder to focus on portions of the target parse tree that correspond with the current time step. As such, we encode $p_2$ using a (unidirectional) LSTM and compute $\boldsymbol{z}_t$ with an attention-weighted average of the LSTM's encoded states at every time step. This attention mechanism is conditioned on the decoder's previous hidden state $\boldsymbol{h}_{t-1}$.

## 3.2 From parse templates to full parses

As mentioned in Section 2.2.1, user-friendly systems should be able to accept high-level parse templates as input rather than full parses. Preliminary experiments show that SCPN struggles to maintain the semantics of the input sentence when we replace the full target parse with templates, and frequently generates short, formulaic sentences. The paraphrase generation model seems to rely heavily on the full syntactic parse to determine output length and clausal ordering, making it difficult to see how to modify the SCPN architecture for template-only target specification.

Instead, we train another model with exactly the same architecture as SCPN to generate complete parses from parse templates. This allows us to do the prediction in two steps: first predict the full syntactic tree and then use that tree to produce the paraphrase. Concretely, for the first step, assume $t_2$ is the parse template formed from the top two levels of the target parse $p_2$. The input to this *parse generator* is the input parse $p_1$ and $t_2$, and it is trained to produce $p_2$. We train the parse generator separately from SCPN (i.e., no joint optimization) for efficiency purposes. At test time, a user only has to specify an input sentence and target template; the template is fed through the parse generator, and its predicted target parse is in turn sent to SCPN for paraphrase generation (see Figure 2).

## 3.3 Template selection and post-processing

By switching from full parses to templates, we have reduced but not completely removed the burden of coming up with a target syntactic form. Certain templates may be not be appropriate for particular input sentences (e.g., turning a long sentence with multiple clauses into a noun phrase). However, others may be too similar to the input syntax, resulting in very little change. Since template selection is not a major focus of this paper, we use a relatively simple procedure, selecting the twenty most frequent templates in PARANMT-

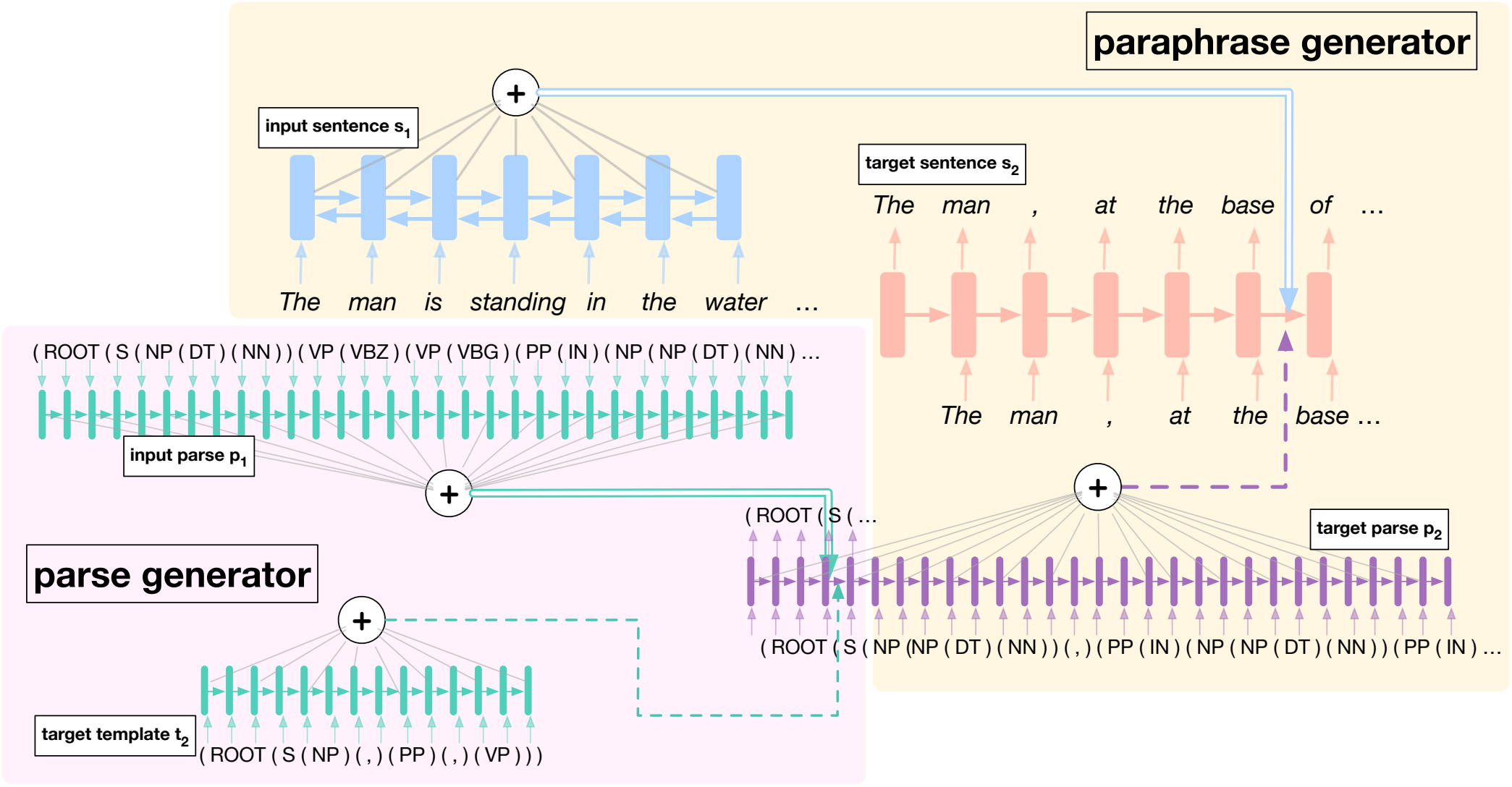

Figure 2: SCPN implements parse generation from templates as well as paraphrase generation from full parses as encoder-decoder architectures (attention depicted with dotted lines, copy mechanism with double stroked lines). While both components are trained separately, at test-time they form a pipelined approach to produce a controlled paraphrase from an input sentence $s_1$, its corresponding parse $p_1$, and a target template $t_2$.

50M.[5]

Since we cannot generate a valid paraphrase for every template, we postprocess to remove nonsensical outputs. In particular, we filter generated paraphrases using n-gram overlap and paraphrastic similarity, the latter of which is computed using the pretrained WORD,TRIAVG sentence embedding model from Wieting and Gimpel (2017).[6] These paraphrastic sentence embeddings significantly outperform prior work due to the PARANMT-50M data.

## 4 Intrinsic Experiments

Before using SCPN to generate adversarial examples on downstream datasets, we need to make sure that its output paraphrases are valid and grammatical and that its outputs follow the specified target syntax. In this section, we compare SCPN to a neural backtranslation baseline (NMT-BT) on the development set of our PARANMT-50M split using both human and automated experiments. NMT-BT is the same pretrained Czech-English model used to create PARANMT-50M; however, here we use it to generate in both directions (i.e., English-Czech and Czech-English).

[5]However, we do provide some qualitative examples of rare and medium-frequency templates in Table 3.

[6]After qualitatively analyzing the impact of different filtering choices, we set minimum n-gram overlap to 0.5 and minimum paraphrastic similarity to 0.7.

| Model | **2** | **1** | **0** |
|---|---|---|---|
| SCPN w/ full parses | 63.7 | 14.0 | 22.3 |
| SCPN w/ templates | 62.3 | 19.3 | 18.3 |
| NMT-BT | 65.0 | 17.3 | 17.7 |

Table 1: A crowdsourced paraphrase evaluation on a three-point scale (**0** = no paraphrase, **1** = ungrammatical paraphrase, **2** = grammatical paraphrase) shows both that NMT-BT and SCPN produce mostly grammatical paraphrases. Feeding parse templates to SCPN instead of full parses does not impact its quality.

### 4.1 Paraphrase quality & grammaticality

To measure paraphrase quality and grammaticality, we perform a crowdsourced experiment in which workers are asked to rate a paraphrase pair $\langle s, g\rangle$ on the three-point scale of Kok and Brockett (2010), where $s$ is the source sentence and $g$ is the generated sentence. A **0** on this scale indicates no paraphrase relationship, while **1** means that $g$ is an ungrammatical paraphrase of $s$ and **2** means that $g$ is a grammatical paraphrase of $s$. We select 100 paraphrase pairs from the development set of our PARANMT-50M split (after the postprocessing steps detailed in Section 3.3) and have three workers rate each pair.[7] To focus the evaluation on the effect of syntactic manipulation on quality, we only select sentences whose top-level parse templates differ (i.e., $t_s \neq t_g$), ensuring that the output of both systems varies syntactically from the source sentences.

[7]We use the Crowdflower platform for our experiments.

The results (Table 1) show that the uncontrolled NMT-BT model's outputs are comparable in quality and grammaticality to those of SCPN; neither system has a significant edge. More interestingly, we observe no quality drop when feeding templates to SCPN (via the parse generator as described in Section 3.2) instead of complete parse trees, which suggests that the parse generator is doing a good job of generating plausible parse trees; thus, for all of the adversarial evaluations that follow, we only use the templated variant of SCPN.

### 4.2 Do the paraphrases follow the target specification?

We next determine how often SCPN's generated paraphrases conform to the target syntax: if $g$ is a generated paraphrase and $p_g$ is its parse, how often does $p_g$ match the ground-truth target parse $p_2$? We evaluate on our development set using *exact template match*: $g$ is deemed a syntactic match to $s_2$ only if the top two levels of its parse $p_g$ matches those of $p_2$. We evaluate two SCPN configurations, where one is given the full target parse $p_2$ and the other is given the result of running our parse generator on the target template $t_2$. As a sanity check, we also evaluate our parse generator using the same metric.

The results (Table 2) show that SCPN does indeed achieve syntactic control over the majority of its inputs. Our parse generator produces full parses that almost always match the target template; however, paraphrases generated using these parses are less syntactically accurate.[8] A qualitative inspection of the generated parses reveals that they can differ from the ground-truth target parse in terms of ordering or existence of lower-level constituents (Table 6); we theorize that these differences may throw off SCPN's decoder.

The NMT-BT system produces paraphrases that tend to be syntactically very similar to the input sentences: 28.7% of these paraphrases have the same template as that of the input sentence $s_1$, while only 11.1% have the same template as the ground-truth target $s_2$. Even though we train SCPN on data generated by NMT backtranslation, we avoid this issue by incorporating syntax into our learning process.

[8]With that said, exact match is a harsh metric; these paraphrases are more accurate than the table suggests, as often they differ by only a single constituent.

| Model | Parse Acc. |
|---|---|
| SCPN w/ gold parse | 64.5 |
| SCPN w/ generated parse | 51.6 |
| Parse generator | 99.9 |

Table 2: The majority of paraphrases generated by SCPN conform to the target syntax, but the level of syntactic control decreases when using generated target parses instead of gold parses. Accuracy is measured by exact template match (i.e., how often do the top two levels of the parses match).

## 5 Adversarial example generation

The intrinsic evaluations show that SCPN produces paraphrases of comparable quality to the uncontrolled NMT-BT system while also adhering to the specified target specifications. Next, we examine the utility of controlled paraphrases for adversarial example generation. To formalize the problem, assume a pretrained model for some downstream task produces prediction $y_x$ given test-time instance $x$. An *adversarial* example $x'$ can be formed by making label-preserving modifications to $x$ such that $y_x \neq y_{x'}$. Our results demonstrate that controlled paraphrase generation with appropriate template selection produces far more valid adversarial examples than backtranslation on sentiment analysis and entailment tasks.

### 5.1 Experimental setup

We evaluate our syntactically adversarial paraphrases on the Stanford Sentiment Treebank (Socher et al., 2013, SST) and SICK entailment detection (Marelli et al., 2014). While both are relatively small datasets, we select them because they offer different experimental conditions: SST contains complicated sentences with high syntactic variance, while SICK almost exclusively consists of short, simple sentences. As a baseline, we compare the ten most probable beams from NMT-BT to controlled paraphrases generated by SCPN using ten templates randomly sampled from the template set described in Section 3.3.[9] We also need pretrained models for which to generate adversarial examples; we use the bidirectional LSTM baseline for both SST and SICK outlined in Tai et al. (2015) since it is a relatively simple architecture that has proven to work well for a variety of problems.[10] Since the SICK task involves characterizing the relationship between two sentences, for simplicity we only generate adversarial examples for the first sentence and keep the second sentence fixed to the ground truth.

[9]We also experimented with the diverse beam search modification proposed by Li et al. (2016b) for NMT-BT but found that it dramatically warped the semantics of many beams; crowdsourced workers rated 49% of its outputs as **0**

| template | paraphrase |
|---|---|
| original | with the help of captain picard , the borg will be prepared for everything . |
| (SBARQ(ADVP)(,)(S)(,)(SQ)) | now , the borg will be prepared by picard , will it ? |
| (S(NP)(ADVP)(VP)) | the borg here will be prepared for everything . |
| (S(S)(,)(CC)(S) (:)(FRAG)) | with the help of captain picard , the borg will be prepared , and the borg will be prepared for everything ... for everything . |
| (FRAG(INTJ)(,)(S)(,)(NP)) | oh , come on captain picard , the borg line for everything . |
| original | you seem to be an excellent burglar when the time comes . |
| (S(SBAR)(,)(NP)(VP)) | when the time comes , you 'll be a great thief . |
| (S(``)(UCP)('')(NP)(VP)) | " you seem to be a great burglar , when the time comes . " you said . |
| (SQ(MD)(SBARQ)) | can i get a good burglar when the time comes ? |
| (S(NP)(IN)(NP)(NP)(VP) | look at the time the thief comes . |

Table 3: Syntactically controlled paraphrases generated by SCPN for two examples from the PARANMT-50M development set. For each input sentence, we show the outputs of four different templates; the fourth template is a failure case (highlighted in green) exhibiting semantic divergence and/or ungrammaticality, which occurs when the target template is unsuited for the input.

### 5.2 Breaking pretrained models

For each dataset, we generate paraphrases for held-out examples and then run a pretrained model over them.[11] We consider a development example $x$ *broken* if the original prediction $y_x$ is correct, but the prediction $y_{x'}$ for at least one paraphrase $x'$ is incorrect. For SST, we evaluate on the binary sentiment classification task and ignore all phrase-level labels (because our paraphrase models are trained on only sentences). Table 4 shows that for both datasets, SCPN breaks many more examples than NMT-BT. Moreover, as shown in Table 5, NMT-BT's paraphrases differ from the original example mainly by lexical substitutions, while SCPN often produces dramatically different syntactic structures.

### 5.3 Are the adversarial examples valid?

We have shown that we can break pretrained models with controlled paraphrases, but are these paraphrases actually valid adversarial examples? After all, it is possible that the syntactic modifications cause informative clauses or words (e.g., negations) to go missing. To measure the validity of our adversarial examples, we turn again to crowdsourced experiments. We ask workers to choose the appropriate label for a given sentence or sentence pair (e.g., positive or negative for SST), and then we compare the worker's judgment to the original development example's label. For both models, we randomly select 100 adversarial examples and have three workers annotate each one. The results (Table 4) show that on the more complex SST data, a higher percentage of SCPN's paraphrases are valid adversarial examples than those of NMT-BT, which is especially encouraging given our model also generates significantly *more* adversarial examples.

### 5.4 Increasing robustness to adversarial examples

If we additionally augment the *training* data of both tasks with controlled paraphrases, we can increase a downstream model's robustness to adversarial examples in the development set. To quantify this effect, we generate controlled paraphrases for the training sets of SST and SICK using the same templates as in the previous experiments. Then, we include these paraphrases as additional training examples and retrain our biLSTM task models.[12] As shown by Table 4, training on SCPN's paraphrases significantly improves robustness to syntactic adversaries without affecting accuracy on the original test sets. One im-on the three-point scale.

[10]We initialize both models using pretrained GloVe embeddings (Pennington et al., 2014) and set the LSTM hidden dimensionality to 300.

[11]Since the SICK development dataset is tiny, we additionally generate adversarial examples on its test set.

[12]We did not experiment with more complex augmentation methods (e.g., downweighting the contribution of paraphrased training examples to the loss).

| | | | No augmentation | | With augmentation | |
|---|---|---|---|---|---|---|
| Model | Task | Validity | Test Acc | Dev Broken | Test Acc | Dev Broken |
| SCPN | SST | 77.1 | 83.1 | 41.8 | 83.0 | 31.4 |
| NMT-BT | SST | 68.1 | 83.1 | 20.2 | 82.3 | 20.0 |
| SCPN | SICK | 77.7 | 82.1 | 33.8 | 82.7 | 19.8 |
| NMT-BT | SICK | 81.0 | 82.1 | 20.4 | 82.0 | 11.2 |

Table 4: SCPN generates more legitimate adversarial examples than NMT-BT, shown by the results of a crowd-sourced validity experiment and the percentage of held-out examples that are broken through paraphrasing. Furthermore, we show that by augmenting the training dataset with syntactically-diverse paraphrases, we can improve the robustness of downstream models to syntactic adversaries (see "Dev Broken" before and after augmentation) without harming accuracy on the original test set.

portant caveat is that this experiment only shows robustness to the set of templates used by SCPN; in real-world applications, careful template selection based on the downstream task, along with using a larger set of templates, is likely to increase robustness to less constrained syntactic adversaries. Augmentation with NMT-BT's paraphrases increases robustness on SICK, but on SST, it degrades test accuracy without any significant gain in robustness; this is likely due to its lack of syntactic variation compared to SCPN.

## 6 Qualitative Analysis

In the previous section, we quantitatively evaluated the SCPN's ability to produce valid paraphrases and adversarial examples. Here, we take a look at actual sentences generated by the model. In addition to analyzing SCPN's strengths and weaknesses compared to NMT-BT, we examine the differences between paraphrases generated by various configurations of the model to determine the impact of each major design decision (e.g., templates instead of full parses).

**Syntactic manipulation:** Table 3 demonstrates SCPN's ability to perform syntactic manipulation, showing paraphrases for two sentences generated using different templates. Many of the examples exhibit complex transformations while preserving both the input semantics and grammaticality, even when the target syntax is very different from that of the source (e.g., when converting a declarative to question). However, the failure cases demonstrate that not every template results in a valid paraphrase, as nonsensical outputs are sometimes generated when trying to squeeze the input semantics into an unsuitable target form.

**Adversarial examples:** Table 5 shows that SCPN and NMT-BT differ fundamentally in the type of adversaries they generate. While SCPN mostly avoids lexical substitution in favor of making syntactic changes, NMT-BT does the opposite. These examples reinforce the results of the experiment in Section 4.2, which demonstrates NMT-BT's tendency to stick to the input syntax. While SCPN is able to break more validation examples than NMT-BT, it is alarming that even simple lexical substitution can break such a high percentage of both datasets we tested.

Ebrahimi et al. (2017) observe a similar phenomenon with HotFlip, their gradient-based substitution method for generating adversarial examples. While NMT-BT does not receive signal from the downstream task like HotFlip, it also does not require external constraints to maintain grammaticality and limit semantic divergence. As future work, it would be interesting to provide this downstream signal to both NMT-BT and SCPN; for the latter, perhaps this signal could guide the template selection process, which is currently fixed to a small, finite set.

**Templates vs. gold parses:** Why does the level of syntactic control decrease when we feed SCPN parses generated from templates instead of gold parses (Table 2)? The first two examples in Table 6 demonstrate issues with the templated approach. In the first example, the template is not expressive enough for the parse generator to produce slots for the highlighted clause. A potential way to combat this type of issue is to dynamically define templates based on factors such as the length of the input sentence. In the second example, a parsing error results in an inaccurate template which in turn causes SCPN to generate a semantically-divergent paraphrase. The final two examples show instances where the templated model performs equally as well as the model with gold parses, displaying the capabilities of our parse generator.

| template | original | paraphrase |
|---|---|---|
| (S(ADVP)(NP)(VP)) | moody , heartbreaking , and filmed in a natural , unforced style that makes its characters seem entirely convincing even when its script is not . | so he 's filmed in a natural , unforced style that makes his characters seem convincing when his script is not . |
| (S(PP)(,)(NP)(VP)) | there is no pleasure in watching a child suffer . | in watching the child suffer , there is no pleasure . |
| (S(S)(,)(CC)(S)) | the characters are interesting and often very creatively constructed from figure to backstory . | the characters are interesting , and they are often built from memory to backstory . |
| | every nanosecond of the the new guy reminds you that you could be doing something else far more pleasurable . | each nanosecond from the new guy reminds you that you could do something else much more enjoyable . |
| | harris commands the screen , using his frailty to suggest the ravages of a life of corruption and ruthlessness . | harris commands the screen , using his weakness to suggest the ravages of life of corruption and recklessness . |

Table 5: Adversarial sentiment examples generated by SCPN (top) and NMT-BT (bottom). The predictions of a pretrained model on the original sentences are correct (red is negative, blue is positive), while the predictions on the paraphrases are incorrect. The syntactically controlled paraphrases of SCPN feature more syntactic modification and less lexical substitution than NMT-BT's backtranslated outputs.

**Removing syntactic control:** To examine the differences between syntactically controlled and uncontrolled paraphrase generation systems, we train an SCPN without including $z_t$, the attention-weighted average of the encoded parse, in the decoder input. This uncontrolled configuration produces outputs that are very similar to its inputs, often identical syntactically with minor lexical substitution. Concretely, the uncontrolled SCPN produces a paraphrase with the same template as its input 38.6% of the time, compared to NMT-BT's 28.7% (Section 4.2).[13]

## 7 Related Work

Paraphrase generation (Androutsopoulos and Malakasiotis, 2010; Madnani and Dorr, 2010) has been tackled using many different methods, including those based on hand-crafted rules (McKeown, 1983), synonym substitution (Bolshakov and Gelbukh, 2004), machine translation (Quirk et al., 2004), and, most recently, deep learning (Prakash et al., 2016; Mallinson et al., 2017; Dong et al., 2017). Our syntactically controlled setting also relates to controlled language generation tasks in which one desires to generate or rewrite a sentence with particular characteristics. We review related work in both paraphrase generation and controlled language generation below.

### 7.1 Data-driven paraphrase generation

Madnani and Dorr (2010) review data-driven methods for paraphrase generation, noting two primary families: template-based and translation-based. The first family includes approaches that use hand-crafted rules (McKeown, 1983), thesaurus-based substitution (Bolshakov and Gelbukh, 2004; Zhang and LeCun, 2015), lattice matching (Barzilay and Lee, 2003), and template-based "shake & bake" paraphrasing (Carl et al., 2005). These methods often yield grammatical outputs but they can be limited in diversity.

The second family includes methods that rewrite the input using methods based on parallel text (Bannard and Callison-Burch, 2005), machine translation (Quirk et al., 2004; Napoles et al., 2016; Suzuki et al., 2017), or related statistical techniques (Zhao et al., 2009). Of particular relevance to our work are methods that incorporate syntax to improve fluency of paraphrase output. Callison-Burch (2008) constrains paraphrases to be the same syntactic type as the input, though he was focused on phrase-level, not sentential, paraphrasing. Pang et al. (2003) learn finite-state automata from translation pairs that generate syntactic paraphrases, though this requires multiple translations into the same language and cannot be used to generate paraphrases outside this dataset. Shen et al. (2006) extend this to deeper syntactic analysis. All of these approaches use syntax to improve grammaticality, which is handled by our decoder language model.

[13] A configuration without the copy mechanism copies input syntax even more, with a 47.7% exact template match.

| | |
|---|---|
| **template** | `(S(CC)(S)(,)(NP)(ADVP)(VP))` |
| **original** | damian encouraged me , criticized , he ... he always made me go a little deeper . |
| **SCPN parse** | but damian , he supported me , he told me , he always made me go a little deeper . |
| **SCPN template** | but damian supported me , he always made me go a little deeper . |
| **template** | `(S(S)(,)(NP)(VP))` |
| **original** | zacharias did n't deserve to die , grishanov thought , and he was aware of the huge irony of his situation |
| **SCPN parse** | zacharias did not deserve to die , grishanov told himself , realizing the greatest irony of all . |
| **SCPN template** | zacharias did not deserve to die , he was aware of the great irony of his situation . |
| **template** | `S(S)(,)(S))` |
| **original** | give me some water , my lips are dry , and i shall try to tell you . |
| **SCPN parse** | give me some water , i have just a dry mouth . |
| **SCPN template** | give me some water , my lips are dry . |
| **template** | `(S(NP)(,)(ADVP)(,)(VP))` |
| **original** | in the meantime , the house is weakened , and all its old alliances and deals are thrown into doubt . |
| **SCPN parse** | the house , meanwhile , is weakening , which will be all of its old alliances and business . |
| **SCPN template** | the house , meanwhile , is weakened , and its old alliances and deals are thrown into doubt . |

Table 6: Examples from PARANMT-50M comparing the output of two SCPN configurations, one with gold target parses (SCPN parse) and one with parses generated from templates (SCPN template), where templates are the top two levels of the gold parses. The first two examples demonstrate issues with missing information caused by inexpressive templates and parsing errors, respectively. The remaining examples, in which both configurations produce syntactically similar paraphrases, showcase the ability of the parse generator to produce viable full parses.

Recent efforts involve neural methods. Iyyer et al. (2014) generate paraphrases with dependency tree recursive autoencoders by randomly selecting parse trees at test time. Li et al. (2017) generate paraphrases using deep reinforcement learning. Gupta et al. (2017) use variational autoencoders to generate multiple paraphrases. These methods differ from our approach in that none offer fine-grained control over the syntactic form of the paraphrase.

### 7.2 Controlled language generation

There is growing interest in generating language with the ability to influence the topic, style, or other properties of the output.

Most related to our methods are those based on syntactic transformations, like the tree-to-tree sentence simplification method of Woodsend and Lapata (2011) based on quasi-synchronous grammar (Smith and Eisner, 2006). Our method is more general since we do not require a grammar and there are only soft constraints. Perhaps the closest to the proposed method is the conditioned recurrent language model of Ficler and Goldberg (2017), which produces language with user-selected properties such as sentence length and formality but is incapable of generating paraphrases.

For machine translation output, Niu et al. (2017) control the level of formality while Sennrich et al. (2016) control the level of politeness. For dialogue, Li et al. (2016a) affect the output using speaker identity, while Wang et al. (2017) develop models to influence topic and style of the output. Shen et al. (2017) perform style transfer on non-parallel texts, while Guu et al. (2017) generate novel sentences from prototypes; again, these methods are not necessarily seeking to generate meaning-preserving paraphrases, merely transformed sentences that have an altered style.

## 8 Conclusion

We propose SCPN, an encoder-decoder model for syntactically controlled paraphrase generation, and show that it is an effective way of generating adversarial examples. Using a parser, we label syntactic variation in large backtranslated data, which provides training data for SCPN. The model exhibits far less lexical variation than existing uncontrolled paraphrase generation systems, instead preferring purely syntactic modifications. It is capable of generating adversarial examples that fool pretrained NLP models. Furthermore, by training on such examples, we increase the robustness of these models to syntactic variation.

## Acknowledgments

We thank the reviewers for their insightful comments. We would also like to thank Mark Yatskar for many useful suggestions on our experiments.